\documentclass{bmvc2k}
\usepackage{booktabs}

\title{Trident Pyramid Networks for Object Detection}

\addauthor{Cédric Picron}{cedric.picron@esat.kuleuven.be}{1}
\addauthor{Tinne Tuytelaars}{tinne.tuytelaars@esat.kuleuven.be}{1}

\addinstitution{ESAT-PSI,\\ KU Leuven,\\ Leuven, Belgium}

\runninghead{Picron et al.}{Trident Pyramid Networks for Object Detection}

\def\eg{\emph{e.g}\bmvaOneDot}

\def\ie{\emph{i.e}\bmvaOneDot}
\begin{document}

\maketitle

\begin{abstract}
    Feature pyramids have become ubiquitous in multi-scale computer vision tasks such as object detection. Given their importance, a computer vision network can be divided into three parts: a backbone (generating a feature pyramid), a neck (refining the feature pyramid) and a head (generating the final output). Many existing networks operating on feature pyramids, named necks, are shallow and mostly focus on communication-based processing in the form of top-down and bottom-up operations. We present a new neck architecture called Trident Pyramid Network (TPN), that allows for a deeper design and for a better balance between communication-based processing and self-processing. We show consistent improvements when using our TPN neck on the COCO object detection benchmark, outperforming the popular BiFPN baseline by $0.5$ AP, both when using the ResNet-50 and the ResNeXt-101-DCN backbone. Additionally, we empirically show that it is more beneficial to put additional computation into the TPN neck, rather than into the backbone, by outperforming a ResNet-101+FPN baseline with our ResNet-50+TPN network by $1.7$ AP, while operating under similar computation budgets. This emphasizes the importance of performing computation at the feature pyramid level in modern-day object detection systems. Code is available at \url{https://github.com/CedricPicron/TPN}.
\end{abstract}

\section{Introduction}
Many computer vision tasks such as object detection and instance segmentation require strong features both at low and high resolution to detect both large and small objects respectively. This is in contrast to the image classification task where low resolution features are sufficient as usually only a single object is present in the center of the image. Networks developed specifically for the image classification task (\eg \cite{simonyan2014very, he2016deep, xie2017aggregated}), further denoted by \textit{backbones}, are therefore insufficient for multi-scale vision tasks. Especially poor performance is to be expected on small objects, as shown in \cite{lin2017feature}.

\begin{figure}
    \vspace{0.1cm}
    \centering
    \includegraphics[scale=0.7]{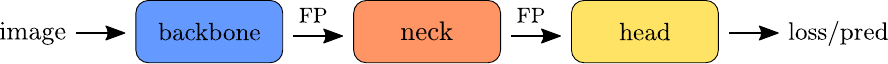}
    \caption{High-level view of a computer vision network. The backbone (\textit{left}) processes the image to output a set of feature maps (\ie a feature pyramid). The neck (\textit{middle}) takes in a feature pyramid (denoted by FP) and returns an updated feature pyramid. Finally, the task-specific head (\textit{right}) produces the loss during training and makes predictions during inference from the final feature pyramid. In this work, we focus on improving the neck.}
    \label{fig:framework}
    \vspace{-0.4cm}
\end{figure}

In order to alleviate this problem, sometimes named the \textit{feature fusion problem}, top-down mechanisms are added \citep{lin2017feature} to propagate semantically strong information from the low resolution to the high resolution feature maps, with improved performance on small objects as a result. Additionally, bottom-up mechanisms can also be appended \citep{liu2018path} such that the lower resolution maps can benefit from the freshly updated higher resolution maps. These top-down and bottom-up mechanisms can now be grouped into a layer, after which multiple of these layers can be concatenated, as done in \cite{tan2020efficientdet}. This part of a computer vision network is called the \textit{neck}, laying in between the \textit{backbone} and the task-specific \textit{head} (see Figure~\ref{fig:framework}).

The top-down and bottom-up operations can be regarded as \textit{communication}-based processing operating on two feature maps, as opposed to \textit{self}-processing operating on a single feature map. Existing necks such as FPN \citep{lin2017feature}, PANet \citep{liu2018path} and BiFPN \citep{tan2020efficientdet} mostly focus on communication-based processing, as this nicely supplements the backbone merely consisting of self-processing. However, when having multiple communication-based operations in a row, communication tends to saturate (everyone is up to date) and hence becomes superfluous. We argue it is therefore more effective to alternate communication-based processing with sufficient self-processing, allowing the feature maps to come up with new findings to be communicated.

\textit{First contribution:} Based on this observation, we design the Trident Pyramid Network (TPN) neck consisting of sequential top-down and bottom-up operations alternated with parallel self-processing mechanisms. The TPN neck is equipped with hyperparameters controlling the amount of communication-based processing and self-processing. These hyperparameters enable our TPN neck to find a better balance between both types of processing compared to other necks as BiFPN \citep{tan2020efficientdet}, outperforming the latter by $0.5$ AP, both when using the ResNet-50 and the ResNeXt-101-DCN backbone.

\textit{Second contribution:} When having additional compute to improve performance, practitioners typically decide to replace their backbone with a heavier one. A ResNet-50+FPN network for example gets traded with the heavier ResNet-101+FPN network. Yet, one might wonder whether it is not more beneficial to add additional computation into the neck (\ie at the feature pyramid level) by using a ResNet-50+TPN network, rather than into the backbone by using a ResNet-101+FPN network. When comparing both options under similar computational characteristics, we show a $1.7$ AP improvement of the ResNet-50+TPN network over the ResNet-101+FPN network. This empirically shows that it is more beneficial to add additional computation into the neck, highlighting the importance of performing computation at the feature pyramid level in modern-day object detection systems.

\section{Related work}
In order to obtain multi-scale features, early detectors performed predictions on feature maps directly coming from the backbone, such as MS-CNN \citep{cai2016unified} and SSD \citep{liu2016ssd}. As the higher resolution maps from the backbone contain relatively weak semantic information, top-down mechanisms were added to propagate semantically strong information from lower resolution maps back to the higher resolution maps as in FPN \citep{lin2017feature} and TDM \citep{shrivastava2016beyond}. Since, many variants and additions have been proposed: PANet \citep{liu2018path} appends bottom-up connections, M2det \citep{zhao2019m2det} uses a U-shape feature interaction architecture, ZigZagNet \citep{lin2019zigzagnet} adds additional pathways between different levels of the top-down and bottom-up hierarchies, NAS-FPN \citep{ghiasi2019fpn} and Hit-Detector \citep{guo2020hit} use Neural Architecture Search (NAS) to automatically design a feature interaction topology, and BiFPN \citep{tan2020efficientdet} modifies PANet by removing some connections, adding skip connections and using weighted feature map aggregation. All of the above variants focus on improving the communication between the different feature maps. We argue however that to be effective, extra self-processing is needed in between the communication flow.

Not all methods use a feature pyramid to deal with scale variation however. TridentNet \citep{li2019scale} applies parallel branches of convolutional blocks with different dilations on a single feature map to obtain scale-aware features. In DetectoRS \citep{qiao2021detectors}, they combine this idea with feature pyramids, by applying their switchable atrous convolutions (SAC) inside their recursive feature pyramids (RFP). Note that to avoid any name confusion with TridentNet, we call our neck by its abbreviated name TPN as opposed to Trident Pyramid Network.

Our TPN neck is also related to networks typically used in segmentation such as U-Net \citep{ronneberger2015u} and stacked hourglass networks \citep{newell2016stacked}, as these networks also use a combination of top-down, self-processing and bottom-up operations. A major difference of these networks with our TPN neck however, is that they do not operate on a feature pyramid in the sense that lower resolution maps are only generated and used within a single layer (\eg within a single hourglass) and are not shared across layers (\eg across two neighboring hourglasses).

\section{Method}
\subsection{TPN neck architecture}
Generally speaking, the neck receives a feature pyramid as input, and outputs an updated feature pyramid. Here, a feature pyramid is defined as a collection of feature maps, with feature maps defined as a collection of feature vectors (called features) organized in a two-dimensional map. More specifically, feature map~$P_l$ denotes a feature map of level $l$ which is $2^l$ times smaller in width and height compared to the initial image resolution. A popular choice for the feature pyramid \citep{lin2017focal} is to consider feature maps $\{P_3, P_4, P_5, P_6, P_7\}$, which we will use as the default setting throughout our discussions and experiments.

\begin{figure}
    \vspace{0.1cm}
    \centering
    \includegraphics[scale=0.6]{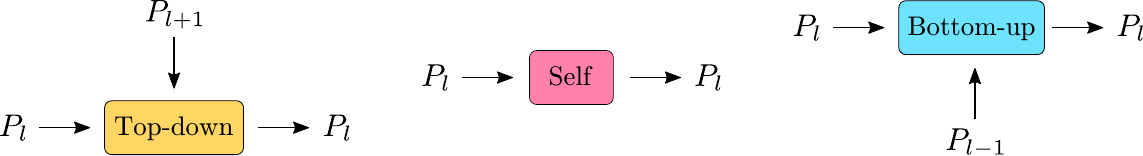}
    \caption{Collection of building blocks for neck architecture design. Here $P_l$ denotes feature map of level~$l$ which is $2^l$ times smaller compared to the initial image resolution. (\textit{Left}) General top-down operation updating feature map $P_l$ with information from lower resolution map $P_{l+1}$. (\textit{Middle}) General self-processing operation updating feature map $P_l$ with information from itself, \ie from feature map $P_l$. (\textit{Right}) General bottom-up operation updating feature map $P_l$ with information from higher resolution map $P_{l-1}$.}
    \label{fig:operations}
    \vspace{-0.4cm}
\end{figure}

The neck is constructed from three building blocks: top-down operations, self-processing operations and bottom-up operations (see Figure~\ref{fig:operations}). In this subsection, we focus on how these operations are best combined, independently of their precise implementations. We call this configuration of operations making up a neck, the \textit{neck architecture}. The specific implementations corresponding to the top-down, self-processing and bottom-up operations will be discussed in Subsection~\ref{sec:self} and Subsection~\ref{sec:td_bu}.

\begin{figure}
    \vspace{0.4cm}
    \centering
    \includegraphics[scale=0.6]{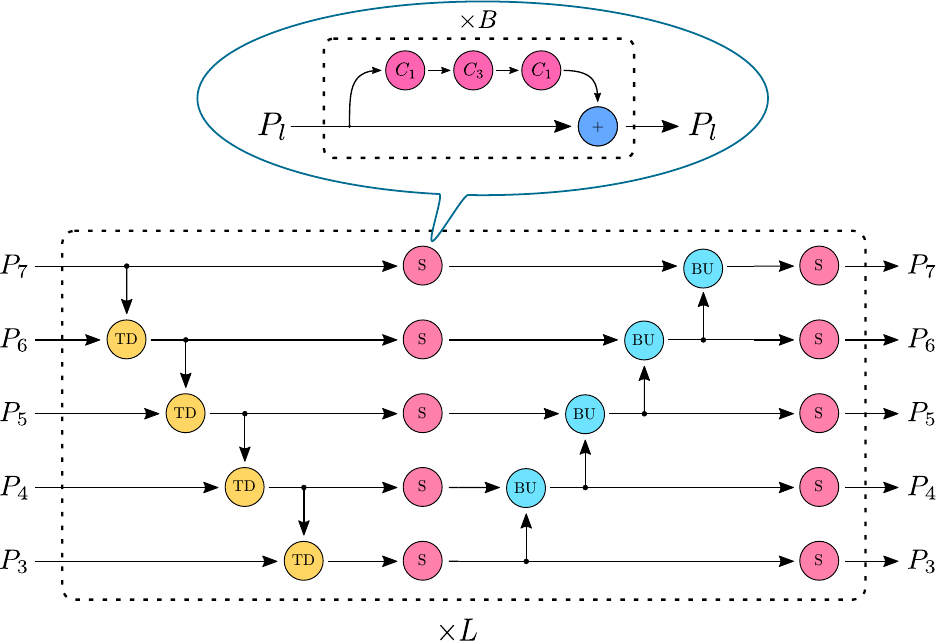}
    \caption{Our TPN neck architecture consisting of $L$ consecutive TPN neck layers (\textit{bottom}), with each self-processing operation consisting of $B$ consecutive bottleneck layers (\textit{top}). The name `Trident Pyramid Network' is inspired by the top-down, first self-processing and bottom-up operations resembling a trident.}
    \label{fig:tpn_detailed}
    \vspace{-0.4cm}
\end{figure}

The TPN neck architecture is displayed in the lower part of Figure~\ref{fig:tpn_detailed} (\ie without text balloon). The TPN neck consists of $L$ consecutive TPN layers, with each layer consisting of sequential top-down and bottom-up operations, alternated by parallel self-processing operations. The TPN neck architecture in itself does not differ much from those of existing necks such as PANet \citep{liu2018path} or BiFPN \citep{tan2020efficientdet}. However, our TPN neck architecture explicitly incorporates self-processing operations which could be of any kind (\eg convolutional or attention-based) and depth (\ie one layer or more). This is in contrast with existing necks as PANet and BiFPN, that use a single convolution layer after each top-down and bottom-up operation, which is considered as a fixed extension of both operations. By doing so, these existing necks are unable to balance the amount of communication-based processing and self-processing, a balance which turns out to be sub-optimal in their case (see Subsection~\ref{sec:balance}). The TPN neck removes this limitation by considering abstract self-processing operations, which could be of any depth, allowing the optimal balance to be found between communication-based processing and self-processing.

\subsection{Self-processing operation} \label{sec:self}
\begin{figure}
    \vspace{0.1cm}
    \centering
    \includegraphics[scale=0.35]{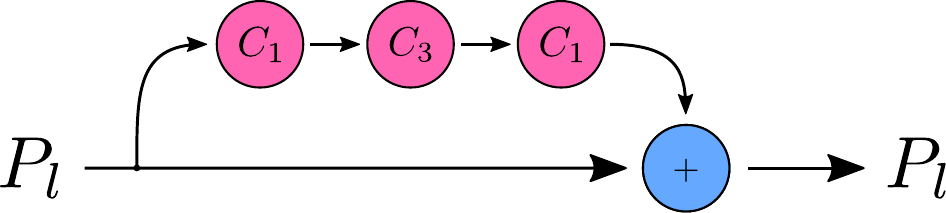}
    \caption{Bottleneck layer used as base self-processing layer. It is a skip-connection operation with a residual branch consisting of three convolution operations: a convolution operation of kernel size~$1$ reducing the original feature size to the hidden feature size, a content convolution operation of kernel size~$3$ applied on the hidden feature size, and finally a convolution operation of kernel size~$1$ expanding the hidden feature size back to the original feature size. Note that each convolution operation (\ie pink convolution node) consists of the actual convolution preceded \citep{he2016identity} by group normalization \citep{wu2018group} and a ReLU activation function.}
    \label{fig:self_bottleneck}
\end{figure}

In general, we consider the self-processing operation to consist of a sequence of $B$ base self-processing layers, where the base self-processing layer could be any convolution or attention-based operation. In this paper, we chose the bottleneck layer from \cite{he2016deep} as the base self-processing layer (see Figure~\ref{fig:self_bottleneck}).

The TPN neck hence consists of $L$ TPN layers, with each TPN layer consisting of top-down and bottom-up operations alternated by $B$ base self-processing layers (see also Figure~\ref{fig:tpn_detailed}). The hyperparameter pair $(L, B)$ balances the amount of communication-based processing and self-processing, with $L$ determining the amount of communication-based processing and $B$ tuning the amount of self-processing. In what follows, we will also refer to the hyperparameter pair $(L, B)$ as the TPN configuration. In Subsection~\ref{sec:balance}, we empirically find out which TPN configurations work best for various computation budgets.

\subsection{Top-down and bottom-up operations} \label{sec:td_bu}
\begin{figure}
    \centering
    \includegraphics[scale=0.3]{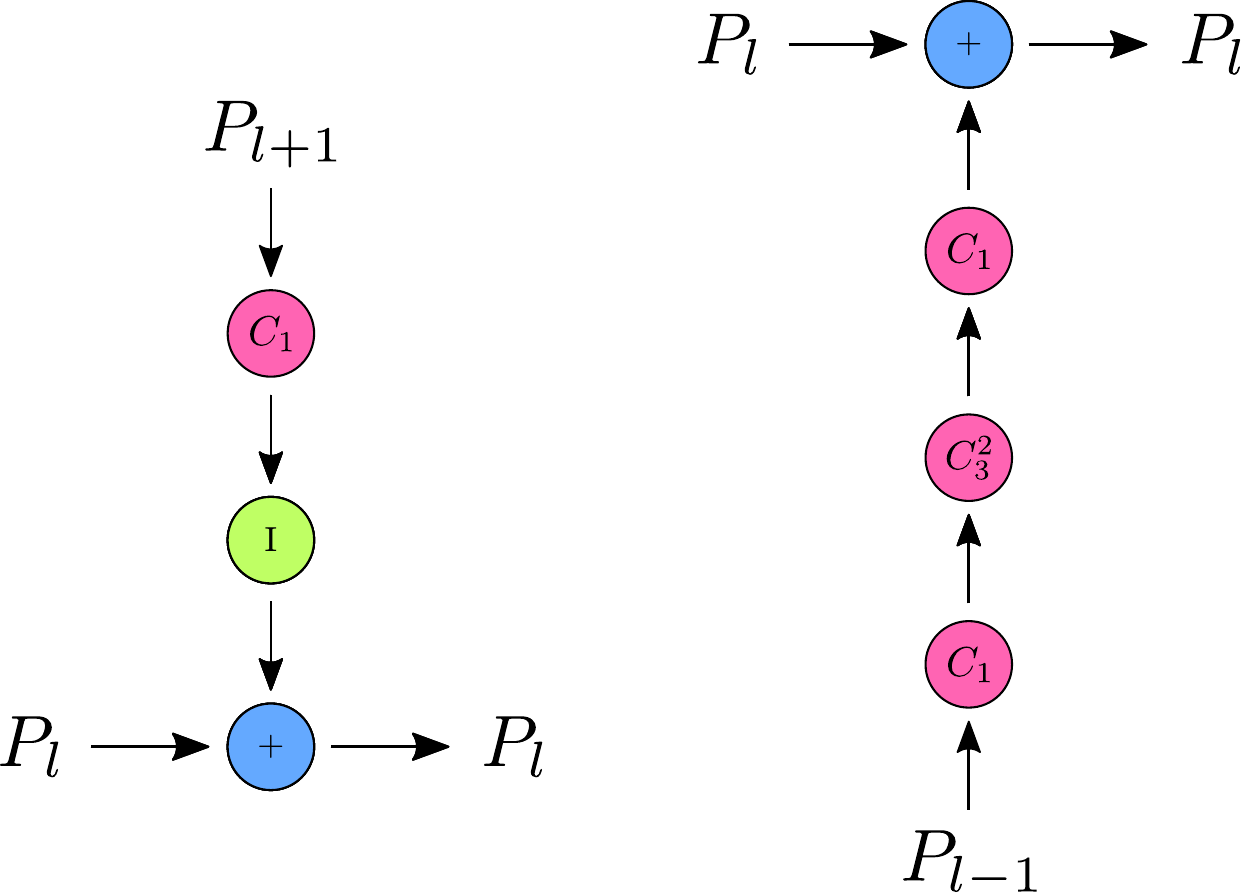}
    \caption{Implementation of the top-down (\textit{left}) and bottom-up (\textit{right}) operations. The pink convolution nodes are defined as in Figure~\ref{fig:self_bottleneck}, with the subscript denoting the kernel size and the superscript denoting the stride (stride~$1$ when omitted). The green node is an interpolation node resizing the input feature map to the required resolution by using bilinear interpolation.}
    \label{fig:def_comp}
\end{figure}

Let us now take a closer look at the top-down and bottom-up operations. Generally speaking, these operations update a feature map based on a second feature map, either having a lower resolution (top-down case) or a higher resolution (bottom-up case). Our implementation of the top-down and bottom-up operations are shown in Figure~\ref{fig:def_comp}. The operations consist of adding a modified version of $P_{l\pm1}$ to $P_l$. This is similar to traditional skip-connection operations, with the exception that the \textit{residual features} originate from a different feature map. The residual branch of the top-down operation consists of a linear projection followed by bilinear interpolation. The presence of the linear projection is important here, as it makes the expectation of the residual features zero at initialization. Failing to do so can be detrimental, especially when building deeper neck modules, as correlated features add up without constraints. An alternative consists in replacing the blue addition nodes with averaging nodes. This however fails to keep the skip connection computation free (due to the $0.5$ factor), which is undesired \citep{he2016identity}. The residual branch of the bottom-up operation is similar to the bottleneck residual branch in Figure~\ref{fig:self_bottleneck}. Only the middle $3 \times 3$ convolution has stride $2$ instead of stride~$1$, avoiding the need for an interpolation step later in the residual branch.

\section{Experiments} \label{sec:experiments}
\subsection{Setup} \label{sec:setup}
\paragraph{Dataset.} 
We perform our experiments on the COCO detection dataset \cite{lin2014microsoft}, where we train on the 2017 COCO training set and evaluate on the 2017 COCO validation or test-dev set.

\paragraph{Implementation details.}
Throughout our experiments, we use ImageNet \citep{deng2009imagenet} pretrained ResNet-50, ResNet-101 and ResNeXt-101-32x4-DCNv2 backbones \citep{he2016deep, xie2017aggregated, zhu2020deformable}, with frozen stem, stage~$1$ and batchnorm layers (see \cite{radosavovic2020designing} for used terminology). 

Our feature pyramid consists of five feature maps, ranging from $P_3$ to $P_7$, each having feature size $256$. Our initial feature pyramid is constructed based on the backbone output feature maps $C_3$ to $C_5$ from stages $2$, $3$ and $4$ respectively. Remember that the subscript denotes how many times the feature map was downsampled with factor $2$ compared to the input image. The initial $P_3$ to $P_5$ maps are obtained by applying simple linear projections on $C_3$ to $C_5$, whereas the initial $P_6$ and $P_7$ maps are obtained by applying a simple network on $C_5$ consisting of $2$ convolutions with stride $2$, with a ReLU activation in between (similar to \cite{lin2017focal}). Throughout our TPN neck modules, we use group normalization \citep{wu2018group} with $8$ groups. For the bottleneck layers (see Figure~\ref{fig:self_bottleneck}), we use a hidden feature size of $64$.

As detection head, we use the one-stage detector head from RetinaNet \citep{lin2017focal}, with $1$ or $4$ hidden layers in both classification and bounding box subnets. We follow the implementation and settings from \cite{wu2019detectron2}, except that the last layer of the subnets has kernel size $1$ (instead of $3$) and that we normalize the losses per feature map (instead of over the whole feature pyramid).

We train our models with the AdamW optimizer \citep{loshchilov2017decoupled} with weight decay $10^{-4}$ using an initial learning rate of $10^{-5}$ for the backbone parameters and an initial learning rate of $10^{-4}$ for the remaining model parameters. We use the same data augmentation scheme as in \cite{carion2020end}. 

Our main experiment results with the ResNet-50 and ResNet-101 backbones in Subsection~\ref{sec:main_experiments}, are obtained by using the 3x training schedule, consisting of $36$ epochs with learning rate drops after the $27$th and $33$rd epoch with a factor $0.1$. For the experiments in Subsection~\ref{sec:balance} and the ResNeXt-101 experiments in Subsection~\ref{sec:large}, we use the 1x training schedule instead, consisting of $12$ epochs with learning rate drops after the $9$th and $11$th epoch with a factor $0.1$. Our ResNet-50 and ResNet-101 models are trained on $2$ GPUs with batch size $2$, whereas our ResNeXt-101 models are trained on a single GPU with batch size $2$. 

\subsection{Main experiments} \label{sec:main_experiments}
\paragraph{Baselines.}
In this subsection, we perform experiments to evaluate the TPN neck. As baseline, we consider the BiFPN neck architecture from \cite{tan2020efficientdet} with batch normalization layers \citep{ioffe2015batch} replaced by group normalization layers \citep{wu2018group}, and with Swish-1 activation functions \citep{ramachandran2017searching} replaced by ReLU activation functions. Multiple of these BiFPN layers will be concatenated such that the BiFPN neck shares similar computational characteristics compared to the tested TPN necks. 

\begin{figure}
    \vspace{0.1cm}
    \centering
    \includegraphics[scale=0.6]{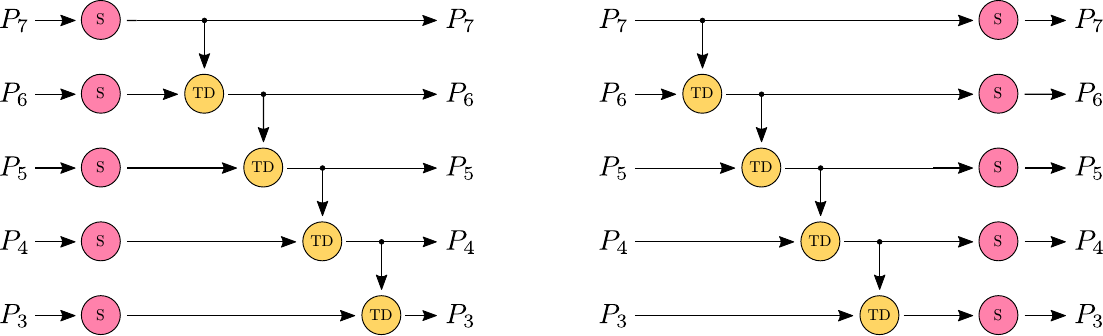}
    \caption{(\textit{Left}) The baseline bFPN neck architecture simulating a heavier backbone followed by a single FPN layer. (\textit{Right}) The baseline hFPN neck architecture simulating a single FPN layer followed by a heavier head.}
    \label{fig:arch_bfpn_hfpn}
    \vspace{-0.4cm}
\end{figure}

Additionally, we would like to compare our TPN neck with the popular FPN neck under similar computation budgets. As the FPN layer was not designed to be concatenated many times, we instead provide the bFPN and hFPN baselines (see Figure~\ref{fig:arch_bfpn_hfpn}). Here the bFPN baseline performs additional self-processing before the FPN simulating a heavier backbone, whereas the hFPN baseline performs additional self-processing after the FPN simulating a heavier head. As such, we will not only be able to evaluate whether the TPN neck outperforms other necks, but also whether it outperforms detection networks using a simple FPN neck with heavier backbones or heads, while operating under similar computation budgets. 

Finally, we compare ResNet-101+FPN and ResNet-101+TPN networks with a ResNet-50+TPN network of similar computation budget, to further assess whether it is more beneficial to put additional computation into the backbone or into the neck.

\paragraph{Results.}
The experiment results evaluating four different TPN configurations and the five baselines, are found in Table~\ref{tab:tpn}.

\begin{table*}
    \centering
    \caption{Experiment results on the 2017 COCO validation set of different TPN necks (top four rows) and its baselines (bottom five rows). The five leftmost columns specify the network (Back = Backbone, Neck, $L$ = Number of layers, $B$ = Number of bottleneck layers per self-processing operation, $C$ = Number of hidden layers in classification and bounding box subnets), the middle six columns show its performance and the five rightmost columns show its computational characteristics. These characteristics were obtained on a GeForce GTX 1660 Ti GPU by applying the network on a batch of two $800 \times 800$ images, each containing $10$ ground-truth objects during training. The training characteristics are found under the columns `Params', `tFPS' and `tMem', whereas the inference characteristics are found under the `iFPS' and `iMem' columns. Here the FPS metric should be interpreted as the number of times the GPU can process above input. Note that both forward and backward passes (with parameter update from the optimizer) are used to obtain the training characteristics. \vspace{0.2cm}}
    \resizebox{\columnwidth}{!}{
        \begin{tabular}{c c c c c|c c c c c c|c c c c c}
            \toprule
            Back & Neck & $L$ & $B$ & $C$ & AP & AP$_{50}$ & AP$_{75}$ & AP$_S$ & AP$_M$ & AP$_L$ & Params & tFPS & tMem & iFPS & iMem \\ \midrule
            R50 & TPN & $1$ & $7$ & $1$ & $41.3$ & $60.5$ & $44.2$ & $26.3$ & $45.9$ & $52.5$ & $36.3$ M & $1.7$ & $3.31$ GB & $5.3$ & $0.50$ GB \\
            R50 & TPN & $2$ & $3$ & $1$ & $41.6$ & $60.9$ & $44.6$ & $26.4$ & $45.8$ & $53.2$ & $36.2$ M & $1.7$ & $3.21$ GB & $5.5$ & $0.50$ GB \\
            R50 & TPN & $3$ & $2$ & $1$ & $\mathbf{41.8}$ & $61.1$ & $44.4$ & $26.2$ & $46.1$ & $53.7$ & $36.7$ M & $1.6$ & $3.27$ GB & $5.3$ & $0.50$ GB \\
            R50 & TPN & $5$ & $1$ & $1$ & $\mathbf{41.8}$ & $\mathbf{61.2}$ & $\mathbf{45.0}$ & $26.0$ & $\mathbf{46.3}$ & $53.4$ & $37.1$ M & $1.6$ & $3.22$ GB & $5.3$ & $0.50$ GB \\
            \midrule
            R50 & BiFPN & $7$ & $-$ & $1$ & $41.3$ & $\mathbf{61.2}$ & $43.7$ & $\mathbf{27.1}$ & $45.2$ & $\mathbf{53.8}$ & $34.7$ M & $1.8$ & $3.22$ GB & $6.0$ & $0.49$ GB \\
            \midrule
            R50 & bFPN & $1$ & $14$ & $1$ & $39.6$ & $60.3$ & $42.4$ & $24.2$ & $43.5$ & $51.3$ & $36.1$ M & $1.7$ & $3.26$ GB & $5.4$ & $0.49$ GB \\
            R50 & hFPN & $1$ & $14$ & $1$ & $40.0$ & $60.2$ & $43.0$ & $25.6$ & $43.9$ & $51.1$ & $36.1$ M & $1.7$ & $3.26$ GB & $5.4$ & $0.49$ GB \\
            \midrule
            R101 & FPN & $1$ & $-$ & $4$ & $40.1$ & $60.1$ & $42.8$ & $24.0$ & $44.0$ & $52.7$ & $55.1$ M & $1.4$ & $3.20$ GB & $4.1$ & $0.57$ GB \\
            R101 & TPN & $1$ & $2$ & $1$ & $40.9$ & $61.0$ & $44.2$ & $25.0$ & $45.3$ & $52.6$ & $51.7$ M & $1.6$ & $3.20$ GB & $4.6$ & $0.55$ GB \\
            \bottomrule
        \end{tabular}}
    \label{tab:tpn}
\end{table*}

First, notice how the $L$ and $B$ hyperparameters (defining the TPN configuration) are chosen in order to obtain models with similar computational characteristics. Here hyperparameter~$L$ denotes the number of consecutive neck layers, while hyperparameter~$B$ denotes the number of bottleneck layers per self-processing operation (see also Figure~\ref{fig:tpn_detailed}). These similar computational characteristics ensure us that a fair comparison can be made between the different models.

Secondly, we observe that the results between the four different TPN configurations are very similar, all four obtaining between $41.3$ and $41.8$ AP. At first glance, it appears that having more TPN neck layers $L$ is slightly more beneficial than having more bottleneck layers $B$ under similar computational budgets. In Subsection~\ref{sec:balance}, we will further investigate which TPN configurations yield the best accuracy vs. efficiency trade-off at various budgets.

Thirdly, when comparing our TPN necks (top four rows) with the BiFPN neck (fifth row), we observe that one TPN configuration performs on par with the BiFPN neck, whereas the remaining three TPN configurations outperform the BiFPN neck by up to $0.5$ AP. We especially notice improvements on the AP$_{75}$ metric, where all four TPN configurations outperform the BiFPN neck by $0.5$ up to $1.3$ AP$_{75}$. This hence shows that our TPN necks provide more accurate detections than the BiFPN neck. In Subsection~\ref{sec:large}, we provide additional results where we compare the TPN neck with the BiFPN neck when using the large ResNeXt-101-32x4-DCNv2 backbone.

Lastly, when comparing the ResNet-50+TPN networks (top four rows) with the ResNet-50+bFPN, ResNet-50+hFPN, ResNet-101+FPN and ResNet-101+TPN baselines (bottom four rows), we again see that the ResNet-50+TPN networks work best. The best-performing baseline (ResNet-101+TPN) from this category is outperformed by all four ResNet-50+TPN configurations with $0.4$ up to $0.9$ AP. Note that the ResNet-101+TPN baseline has considerably more parameters and is clearly slower at inference, but still does not match the performance of the ResNet-50+TPN networks despite its higher computational cost. This hence shows that the TPN neck not only outperforms other necks such as BiFPN, but also other detection networks using heavier backbones or heads while operating under similar overall computation budgets. This highlights the importance of necks operating on feature pyramids within general object detection networks.

\subsection{Comparison between different TPN configurations} \label{sec:balance}

\begin{figure}
    \centering
    \begin{minipage}{0.329\textwidth}
        \centering
        \includegraphics[scale=0.28]{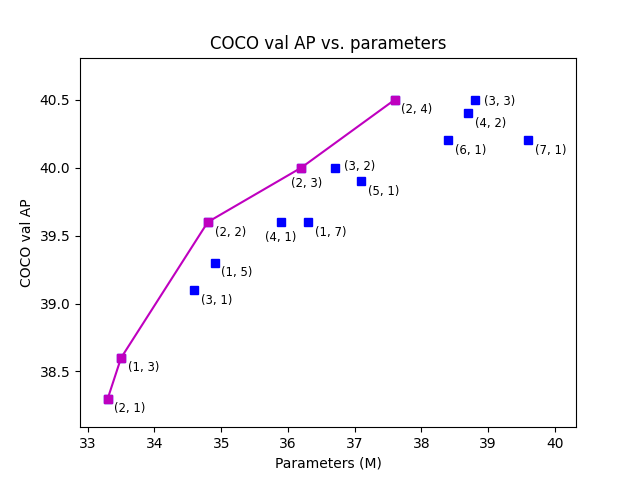}
    \end{minipage} \hfill
    \begin{minipage}{0.329\textwidth}
        \centering
        \includegraphics[scale=0.28]{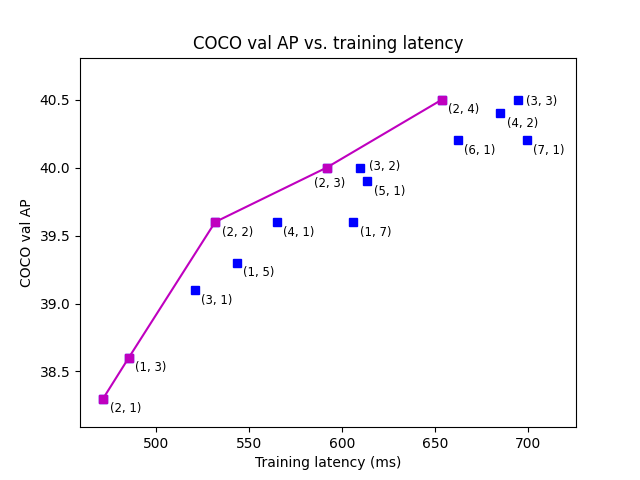}
    \end{minipage} \hfill
    \begin{minipage}{0.329\textwidth}
        \centering
        \includegraphics[scale=0.28]{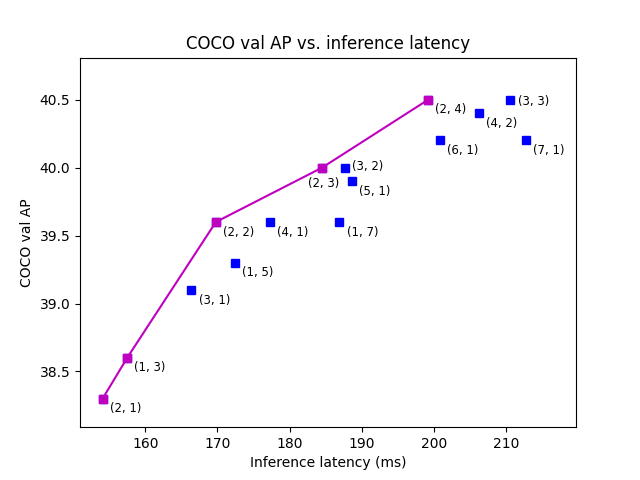}
    \end{minipage} \hspace{-5pt} \vspace{0.1cm}
    \caption{Accuracy vs. efficiency comparisons between $15$ different $(L, B)$ TPN configurations using the `parameters' (\textit{left}), `training latency' (\textit{middle}) and `inference latency' (\textit{right}) metrics. The accuracies correspond to the COCO validation APs, obtained after training the models for $12$ epochs using the 1x schedule. The TPN configurations yielding the best accuracy vs. efficiency trade-off at various computation budgets, are highlighted in magenta.}
    \label{fig:balance}
\end{figure}

In this subsection, we investigate which TPN configurations yield the best accuracy vs. efficiency trade-off at various computation budgets. Here, a TPN configuration is determined by the hyperparameter pair $(L, B)$, respectively denoting the number of TPN layers and the number of bottleneck layers per self-processing operation. In Figure~\ref{fig:balance}, we compare $15$ different $(L, B)$ TPN configurations using the `parameters', `training latency' and `inference latency' metrics, with the latency metrics obtained using the same methodology as in Table~\ref{tab:tpn}.

We can see from the magenta curves yielding the optimal TPN configurations at various computation budgets, that having a good balance between communication-based processing (in the form of TPN layers $L$) and self-processing (in the form of bottleneck layers per self-processing operation $B$) is important. We can for example see that the balanced $(2, 2)$ configuration outperforms the unbalanced $(3, 1)$ and $(1, 5)$ configurations. The same observation can also be made at higher computation budgets, where the balanced $(2, 4)$, $(4, 2)$ and $(3, 3)$ configurations outperform the unbalanced $(6, 1)$ and $(7, 1)$ configurations.

We hence empirically show that balanced $(L, B)$ configurations with $L \geq 2$ and $B \geq 2$ are to be preferred over unbalanced configurations such as $(L, 1)$ and $(1, B)$. By having only one self-processing operation in between each communication-based operation, existing necks with a $(L, 1)$ configuration such as PANet \citep{liu2018path} and BiFPN \citep{tan2020efficientdet} crucially lack self-processing. The TPN neck solves this problem by introducing the $(L, B)$ hyperparameter pair, such that a better balance between communication-based processing and self-processing can be chosen.

\subsection{Comparison between TPN and BiFPN when using large backbone} \label{sec:large}

We provide additional results comparing the TPN neck with the BiFPN neck on the 2017 COCO test-dev set, when using the large ResNeXt-101-32x4-DCNv2 \citep{xie2017aggregated, zhu2020deformable} backbone.

The results are found in Table~\ref{tab:large}. We can see that the TPN neck outperforms the BiFPN neck by $0.5$ AP, while having a similar computation budget. We can moreover see that the TPN neck has a better performance across all object scales. These additional experimental results show that the superiority of the TPN neck compared to the BiFPN neck generalizes to larger backbones and to the 2017 COCO test-dev set.

\begin{table*}
    \centering
    \caption{Experiment results on the 2017 COCO test-dev set comparing the TPN neck (top row) with the BiFPN neck (bottom row), when using the large ResNeXt-101-32x4-DCNv2 backbone. The five leftmost columns specify the model, the six middle columns show the model performance, and the five rightmost columns contain the computational characteristics of the model. These characteristics are obtained as explained in Table~\ref{tab:tpn}, except that we use a batch of two $600 \times 600$ images. \vspace{0.2cm}}
    \resizebox{\columnwidth}{!}{
    \begin{tabular}{c c c c c|c c c c c c|c c c c c}
        \toprule
        Backbone & Core & $B$ & $L$ & $C$ & AP & AP$_{50}$ & AP$_{75}$ & AP$_S$ & AP$_M$ & AP$_L$ & Params & tFPS & tMem & iFPS & iMem \\ \midrule
        X101-DCNv2 & TPN & $2$ & $3$ & $1$ & $44.3$ & $65.4$ & $48.0$ & $27.0$ & $47.6$ & $56.4$ & $59.1$ M & $1.1$ & $3.60$ GB & $4.7$ & $0.39$ GB \\
        X101-DCNv2 & BiFPN & $-$ & $7$ & $1$ & $43.8$ & $64.8$ & $47.3$ & $26.4$ & $47.5$ & $55.3$ & $57.1$ M & $1.2$ & $3.59$ GB & $5.1$ & $0.38$ GB \\
        \bottomrule
    \end{tabular}}
    \label{tab:large}
    \vspace{-0.4cm}
\end{table*}

\section{Conclusion}
In this work, we introduce the TPN neck consisting of top-down, self-processing and bottom-up operations. By considering the self-processing operation as an independent entity separated from the top-down and bottom-up operations, the TPN neck is able to find a better balance between communication-based processing and self-processing compared to existing necks as BiFPN. We validate our findings on the COCO object detection benchmark where we show the superiority of the TPN neck compared to the BiFPN neck. We additionally observe that moving computation from the TPN neck to the backbone or head decreases performance, highlighting the importance and effectiveness of the neck component within object detection networks.

\section*{Acknowledgements}
This work was supported by the KU Leuven C1 MACCHINA project.

\bibliography{egbib}
\end{document}